\documentclass{article}

\usepackage{microtype}
\usepackage{graphicx}
\usepackage{subfigure}
\usepackage{booktabs} % for professional tables
\usepackage{xcolor}
\usepackage{color}
\usepackage{siunitx}

\usepackage{listings}

\definecolor{lightgray}{rgb}{.9,.9,.9}
\definecolor{darkgray}{rgb}{.4,.4,.4}
\definecolor{purple}{rgb}{0.65, 0.12, 0.82}

\lstdefinelanguage{GLSL}{
  keywords={ivec2, $, void, int, float, for, vec4, dot, const},
  keywordstyle=\color{blue},
  ndkeywords={getOutputCoords, getA, getB, setOutput},
  ndkeywordstyle=\color{black}\bfseries,
  identifierstyle=\color{black},
  sensitive=false,
  comment=[l]{//},
  morecomment=[s]{/*}{*/},
  commentstyle=\color{purple}\ttfamily,
  stringstyle=\color{red}\ttfamily,
  morestring=[b]',
  morestring=[b]"
}

\usepackage{hyperref}

\usepackage[accepted]{sysml2019}

\sysmltitlerunning{TensorFlow.js: Machine Learning for the Web and Beyond}

\begin{document}

\twocolumn[
\sysmltitle{TensorFlow.js: Machine Learning for the Web and Beyond}

% It is OKAY to include author information, even for blind
% submissions: the style file will automatically remove it for you
% unless you've provided the [accepted] option to the sysml2019
% package.

% List of affiliations: The first argument should be a (short)
% identifier you will use later to specify author affiliations
% Academic affiliations should list Department, University, City, Region, Country
% Industry affiliations should list Company, City, Region, Country

% You can specify symbols, otherwise they are numbered in order.
% Ideally, you should not use this facility. Affiliations will be numbered
% in order of appearance and this is the preferred way.
\sysmlsetsymbol{equal}{*}

\begin{sysmlauthorlist}
\sysmlauthor{Daniel Smilkov}{equal,goo}
\sysmlauthor{Nikhil Thorat}{equal,goo}
\sysmlauthor{Yannick Assogba}{goo}
\sysmlauthor{Ann Yuan}{goo}
\sysmlauthor{Nick Kreeger}{goo}
\sysmlauthor{Ping Yu}{goo}
\sysmlauthor{Kangyi Zhang}{goo}
\sysmlauthor{Shanqing Cai}{goo}
\sysmlauthor{Eric Nielsen}{goo}
\sysmlauthor{David Soergel}{goo}
\sysmlauthor{Stan Bileschi}{goo}
\sysmlauthor{Michael Terry}{goo}
\sysmlauthor{Charles Nicholson}{goo}
\sysmlauthor{Sandeep N. Gupta}{goo}
\sysmlauthor{Sarah Sirajuddin}{goo}
\sysmlauthor{D. Sculley}{goo}
\sysmlauthor{Rajat Monga}{goo}
\sysmlauthor{Greg Corrado}{goo}
\sysmlauthor{Fernanda B. Vi\'egas}{goo}
\sysmlauthor{Martin Wattenberg}{goo}

\end{sysmlauthorlist}

\sysmlaffiliation{goo}{Google Brain}

\sysmlcorrespondingauthor{Daniel Smilkov}{smilkov@google.com}
\sysmlcorrespondingauthor{Nikhil Thorat}{nsthorat@google.com}

% You may provide any keywords that you
% find helpful for describing your paper; these are used to populate
% the "keywords" metadata in the PDF but will not be shown in the document
\sysmlkeywords{deep learning, library, neural network, javascript, webgl, node.js}

\vskip 0.3in

\begin{abstract}
TensorFlow.js is a library for building and executing machine learning algorithms in JavaScript. TensorFlow.js models run in a web browser and in the Node.js environment. The library is part of the TensorFlow ecosystem, providing a set of APIs that are compatible with those in Python, allowing models to be ported between the Python and JavaScript ecosystems. TensorFlow.js has empowered a new set of developers from the extensive JavaScript community to build and deploy machine learning models and enabled new classes of on-device computation. This paper describes the design, API, and implementation of TensorFlow.js, and highlights some of the impactful use cases.

\end{abstract}
]

% this must go after the closing bracket ] following \twocolumn[ ...

% This command actually creates the footnote in the first column
% listing the affiliations and the copyright notice.
% The command takes one argument, which is text to display at the start of the footnote.
% The \sysmlEqualContribution command is standard text for equal contribution.
% Remove it (just {}) if you do not need this facility.

%\printAffiliationsAndNotice{}  % leave blank if no need to mention equal contribution
\printAffiliationsAndNotice{\sysmlEqualContribution} % otherwise use the standard text.

\section{Introduction}
Machine learning (ML) has become an important tool in software systems, enhancing existing applications and enabling entirely new ones. However, the available software platforms for ML reflect the academic and industrial roots of the technology. Production-quality ML libraries are typically written for Python and C++ developers. Nonetheless, there is a vast community of both frontend and backend JavaScript (JS) developers that continues to grow at a high pace. There were 2.3 million GitHub pull requests in JS in 2017, compared to 1 million in Python \cite{octoverse}. According to the Stack Overflow Developer Survey in 2018, JS is the most commonly used programming language \cite{stackoverflow-survey-2018}.

%and while there have been a few ML JS libraries, little attention has been given to production-quality JS libraries meant to be used in products, work on a wide range of devices, are well-tested, and are extensible.

%The JavaScript community matters; 

This lack of attention matters. The JS environment has the potential to support a new and distinctive class of applications. On-device computation has a number of benefits, including data privacy, accessibility, and low-latency interactive applications. Empowering the community of JS developers may lead to new classes of applications.

This paper describes the design and development of the TensorFlow.js library, which was motivated by the importance of the JS community and web-based applications for ML. A first-class citizen in the TensorFlow \cite{TensorFlow} ecosystem, the platform brings high-performance ML and numeric computation capabilities to JS. While several open source JS platforms for ML have appeared, to our knowledge TensorFlow.js is the first to enable integrated training and inference on the GPU from the browser, and offer full Node.js integration for server-side deployment. We have attempted to ensure that TensorFlow.js meets a high standard of productionization, including high-level libraries, comprehensive testing, and explicit extensibility. It has already seen significant uptake by the JS community.

We discuss three main aspects of our experience building TensorFlow.js. First, we describe some of the unique challenges and advantages of the JS environment. Second, we cover the design details of the library, its APIs, which represents a balance between standard web development practice and compatibility with TensorFlow, and the techniques we used to overcome the limitations of the JS environment. Finally, we describe a few interesting and new use cases that have been enabled by TensorFlow.js.

\section{Background and related work}
\label{sec:background}

The design of TensorFlow.js is grounded in specific constraints of the JS environment. Here we detail the technical challenges of ML with JS and related efforts to address them. 

\subsection{The JavaScript environment}

\textbf{Different environments.} One of the challenges of JS is that it runs in different environments. Computation can happen client-side in a browser, server-side, most notably as part of the Node.js framework, and more recently on the desktop via frameworks like Electron. TensorFlow.js is designed to work in all these settings, although the majority of our work to date has been tuning it for client-side development in a web browser.

\textbf{Performance.} A second key challenge, specific to the browser environment, is performance. JS is an interpreted language so it does not typically match the speed of a compiled language like C++ or Java for numerical computation. Unlike Python which can bind to C++ libraries, browsers do not expose this capability. For security reasons, browser applications don't have direct access to the GPU, which is typically where numerical computation happens for modern deep learning systems.

To address these performance issues, a few new JS standards are emerging. One notable solution is WebAssembly \cite{webassembly}, a method for compiling C++ programs to bytecode that can be interpreted and executed directly in the browser. For certain tasks, WebAssembly can outperform plain JS. Most modern browsers also support WebGL \cite{webgl}, an API that exposes OpenGL to JS. OpenGL is a cross-language, cross-platform API for rendering 2D and 3D vector graphics, enabling games and other high-performance rendering tasks directly in a webpage. On the server-side, JS libraries can bind to existing native modules that are written in C and C++ via Node.js's N-API interface \cite{napi}.

\textbf{Cross-browser compatibility.} JS is designed to be a cross-platform language supported by all major browsers with standardized Web APIs that make it easy to write applications that run on all platforms. In practice, browsers are built by several different vendors with slightly different implementations and priorities. For example, while Chrome and Firefox support WebGL 2.0 (a significant improvement over WebGL 1.0), Apple's Safari has settled on WebGL 1.0 and shifted focus to future technologies such as WebGPU \cite{webgpu}. Web application authors have to work hard to hide this inconsistency in their applications, often requiring extensive testing infrastructure to test across large number of platforms.

\textbf{Single-threaded execution.} One of the other challenges of the JS environment is its single threaded nature. JS has a `main thread' (also known as the `UI thread'), which is where webpage layout, JS code, event processing and more happen. While this greatly simplifies some aspects of the development model, it also means that application developers need to be careful not to block the main thread as it will cause other parts of the page to slow down. A well-designed JS library therefore requires a careful balance between the simplicity of synchronous APIs and the non-blocking benefits of asynchronous APIs.

\subsection{Opportunities in a browser-based environment}

\textbf{Shareability.} A major motivation behind TensorFlow.js is the ability to run ML in standard browsers, without any additional installations. Models and applications written in TensorFlow.js are easily shared on the web, lowering the barrier to entry for machine learning. This is particularly important for educational use cases and for increasing the diversity of contributors to the field.

\textbf{Interactivity.} From a machine learning perspective, the interactive nature of web browsers and versatile capabilities of Web APIs open the possibility for a wide range of novel user-centric ML applications which can serve both education and research purposes. Visualizations of neural networks such as \cite{olah2014} and \cite{playground2016} have been popular to teach the basic concepts of machine learning.

\textbf{On-device computation.} Lastly, standardized access to various components of device hardware such as the web camera, microphone, and the accelerometer in the browser allow easy integration between ML models and sensor data. An important result of this integration is that user data can stay on-device and preserve user-privacy, enabling applications in the medical, accessibility, and personalized ML domains. For example, speech-impaired users can use their phones to collect audio samples to train a personalized model in the browser. Another technology, called Federated Learning \cite{federated-learning}, enables devices to collaboratively train a centralized model while keeping sensitive data on device. Browsers are a natural a platform for this type of application.

\subsection{Related work}
\label{sec:related}
Given the popularity and the unique benefits of the JS ecosystem, it is no surprise that many open-source browser-based ML libraries exist. ConvNetJS \cite{convnetjs}, Synaptic \cite{synaptic}, Brain.js \cite{brainjs}, Mind \cite{mind} and Neataptic \cite{neataptic} each provide a simple JS API that allows beginners to build and train neural networks with only a few lines of code. More specialized JS ML libraries include Compromise \cite{compromise} and Natural \cite{Natural}, which focus on NLP applications, and NeuroJS \cite{neurojs} and REINFORCEjs \cite{reinforcejs}, which focus on reinforcement learning. ML.js \cite{mljs} provides a more general set of ML utilities, similar to the Python-based scikit-learn \cite{scikit-learn}.

% Users interested in a more general set of ML utilities, similar to scikit-learn \cite{scikit-learn} in Python, might find ML.js  useful.

%These libraries do not take advantage of hardware acceleration in the browser, making them less useful in real-world applications and production use-cases.

These libraries do not provide access to hardware acceleration from the browser which we have found to be important for computational efficiency and minimizing latency for interactive use cases and state of the art ML models. A few libraries have attempted to take advantage of hardware acceleration, notably TensorFire \cite{tensorfire}, Propel (built on top of TensorFlow.js) \cite{propel} and Keras.js \cite{kerasjs}, however they are no longer actively maintained.

WebDNN \cite{webdnn} is another deep learning library in JS that can execute pretrained models developed in TensorFlow, Keras, PyTorch, Chainer and Caffe. To accelerate computation, WebDNN uses WebGPU \cite{webgpu}, a technology initially proposed by Apple. WebGPU is in an early exploratory stage and currently only supported in Safari Technology Preview, an experimental version of the Safari browser. As a fallback for other browsers, WebDNN uses WebAssembly \cite{webassembly}, which enables execution of compiled C and C++ code directly in the browser. While WebAssembly has support across all major browsers, it lacks SIMD instructions, a crucial component needed to make it as performant as WebGL and WebGPU.

\section{Design and API}

The goals of TensorFlow.js differ from other popular ML libraries in a few important ways. Most notably, TensorFlow.js was designed to bring ML to the JS ecosystem, empowering a diverse group of JS developers with limited or no ML experience \cite{carrie}. At the same time, we wanted to enable experienced ML users and teaching enthusiasts to easily migrate their work to JS, which necessitated wide functionality and an API that spans multiple levels of abstraction. These two goals are often in conflict, requiring a fine balance between ease-of-use and functionality. Lastly, as a new library with a growing user base, missing functionality was prioritized over performance.

%A major goal of the library was to bring TensorFlow to a large number of JS developers with no or limited ML experience.

These goals differ from popular deep learning libraries \cite{TensorFlow, pytorch}, where performance is usually the number one goal, as well as other JS ML libraries (see Section~\ref{sec:related}), whose focus is on simplicity over completeness of functionality. For example, a major differentiator of TensorFlow.js is the ability to author and train models directly in JS, rather than simply being an execution environment for models authored in Python.

\subsection{Overview}

The API of TensorFlow.js is largely modeled after TensorFlow, with a few exceptions that are specific to the JS environment. Like TensorFlow, the core data structure is the \textit{Tensor}. The TensorFlow.js API provides methods to create tensors from JS arrays, as well as mathematical functions that operate on tensors.

%The TensorFlow.js API also provides methods to retrieve the values to CPU-resident arrays from \code{Tensor}s.

\begin{figure}[hp]
\includegraphics[width=3in]{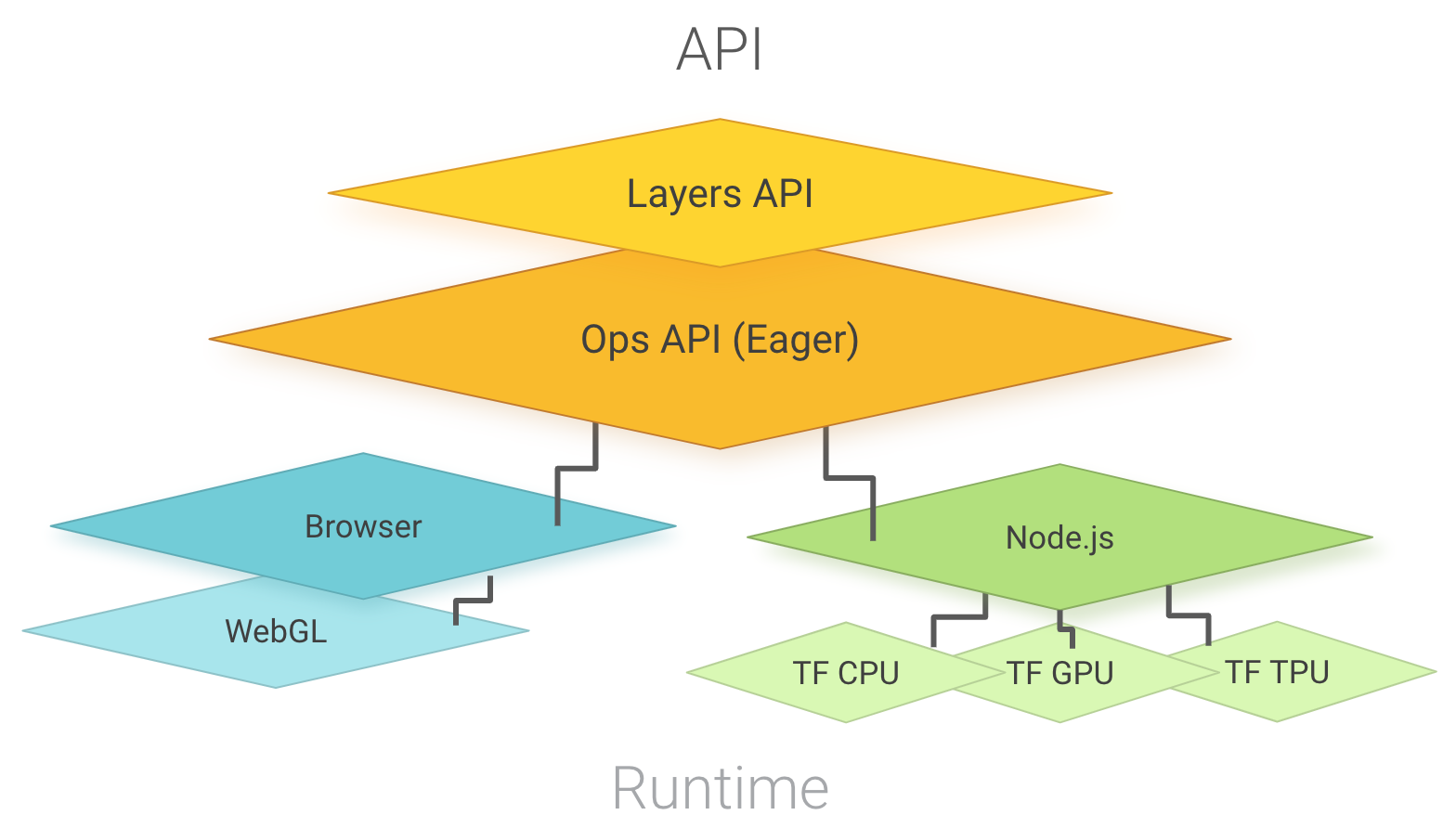}
\caption{Overview of the TensorFlow.js architecture}
\label{fig:overview}
\end{figure}

Figure~\ref{fig:overview} shows a high level schematic view of the architecture. TensorFlow.js consists of two sets of APIs: the \textit{Ops API} which provides lower-level linear algebra operations (e.g. matrix multiplication, tensor addition, etc.), and the \textit{Layers API}, which provides higher-level model building blocks and best practices with emphasis on neural networks. The \textit{Layers API} is modeled after the \textit{tf.keras} namespace in TensorFlow Python, which is based on the widely adopted Keras API \cite{keras}.

TensorFlow.js is designed to run in-browser and server-side, as shown in Figure~\ref{fig:overview}. When running inside the browser, it utilizes the GPU of the device via WebGL to enable fast parallelized floating point computation. In Node.js, TensorFlow.js binds to the TensorFlow C library, enabling full access to TensorFlow.
% Under the hood, the C API can use either the host's CPU (taking advantage of AVX instructions) or the GPU (using the CUDA library \cite{cuda}).
TensorFlow.js also provides a \textit{slower} CPU implementation as a fallback (omitted in the figure for simplicity), implemented in plain JS. This fallback can run in any execution environment and is automatically used when the environment has no access to WebGL or the TensorFlow binary.

\subsection{Layers API}

Beginners and others who are not interested in the operation-level details of their model might find the low-level operations API complex and error prone. The widely adopted Keras library \cite{keras}, on the other hand, provides higher-level building blocks with emphasis on deep learning. With its carefully thought out API, Keras is popular among deep learning beginners and applied ML practitioners.  At the heart of the API is the concept of a model and layers. Users can build a model by assembling a set of pre-defined layers, where each layer has reasonable default parameters to reduce cognitive load.

For these reasons, TensorFlow.js provides the \textit{Layers API}, which mirrors the Keras API as closely as possible, including the serialization format. This enables a two-way door between Keras and TensorFlow.js; users can load a pretrained Keras model (see Section~\ref{sec:converter}) in TensorFlow.js, modify it, serialize it, and load it back in Keras Python.

Listing~\ref{lst:layers-api} shows an example of training a model using the Layers API.

\begin{minipage}{\linewidth} % Disable page breaks
\begin{lstlisting}[caption={An example TensorFlow.js program that shows how to build a single-layer linear model with the layers API, train it with synthetic data, and make a prediction on an unseen data point.},label={lst:layers-api}]
// A linear model with 1 dense layer.
const model = tf.sequential();
model.add(tf.layers.dense({
  units: 1, inputShape: [1]
}));

// Specify the loss and the optimizer.
model.compile({
  loss: 'meanSquaredError',
  optimizer: 'sgd'
});

// Generate synthetic data to train.
const xs =
    tf.tensor2d([1, 2, 3, 4], [4, 1]);
const ys =
    tf.tensor2d([1, 3, 5, 7], [4, 1]);

// Train the model using the data.
model.fit(xs, ys).then(() => {
  // Do inference on an unseen data point
  // and print the result.
  const x = tf.tensor2d([5], [1, 1]);
  model.predict(x).print();
});
\end{lstlisting}
\end{minipage}

\subsection{Operations and Kernels}

As in TensorFlow, an \textit{operation} represents an abstract computation (e.g. matrix multiplication) that is independent of the physical device it runs on.
%The lowest level API that TensorFlow provides to the user is of Tensors and Operations.
Operations call into \textit{kernels}, which are device-specific implementations of mathematical functions which we go over in Section~\ref{sec:impl}.

%In the browser, kernels are WebGL fragment shader programs that read from an input texture and generate another texture. In Node.js with a TensorFlow C backend, kernels call into the TensorFlow C API.

\subsection{Backends}

To support device-specific kernel implementations, TensorFlow.js has a concept of a \textit{Backend}. A backend implements kernels as well as methods such as \textit{read()} and \textit{write()} which are used to store the \textit{TypedArray} that backs the tensor. Tensors are decoupled from the data that backs them, so that operations like reshape and clone are effectively free. This is achieved by making shallow copies of tensors that point to the same data container (the \textit{TypedArray}). When a tensor is disposed, we decrease the reference count to the underlying data container and when there are no remaining references, we dispose the data container itself.

%Fig.~\ref{fig:stack} shows a simplified call stack when the user calls the \textit{tf.matMul()} op and its delegation in two different environments: the browser on the left side, and Node.js on the right side. A call to the operation can result in one or more calls to the underlying kernels that each backend implements.

% \begin{figure}[htbp]
% \includegraphics[width=3.5in]{tfjs-stack.png}
% \caption{The call stack of matrix multiplication in two different environments: the browser (left) and Node.js (right).}
% \label{fig:stack}
% \end{figure}

%For instance, when a kernel is called using the WebGL backend, the backend finds the WebGL texture associated with the tensor's \code{dataId}, executes the WebGL program which writes to a WebGL texture, creates a new \code{dataId} pointing to the output WebGL texture, and returns a tensor that wraps the \code{dataId}.

\subsection{Automatic differentiation}

Since wide functionality was one of our primary design goals, TensorFlow.js supports automatic differentiation, providing an API to train a model and to compute gradients.

The two most common styles of automatic differentiation are graph-based and eager. Graph-based engines provide an API to construct a computation graph, and execute it later. When computing gradients, the engine statically analyzes the graph to create an additional gradient computation graph. This approach is better for performance and lends itself easily to serialization.

Eager differentiation engines, on the other hand, take a different approach \cite{pytorch, TensorFlow, autograd}. In eager mode, the computation happens immediately when an operation is called, making it easier to inspect results by printing or using a debugger. Another benefit is that all the functionality of the host language is available while your model is executing; users can use native \textit{if} and \textit{while} loops instead of specialized control flow APIs that are hard to use and produce convoluted stack traces.

%When computing gradients, an internal tape records all of the operations that were executed during the forward pass. When a gradient is computed, another set of operations representing the gradient are generated from the tape.
Due to these advantages, eager-style differentiation engines, like TensorFlow Eager \cite{tensorflow-eager} and PyTorch \cite{pytorch}, are rapidly gaining popularity. Since an important part of our design goals is to prioritize ease-of-use over performance, TensorFlow.js supports the eager style of differentiation.

\subsection{Asynchronous execution}
\label{sec:async}
JS runs in a single thread, shared with tasks like page layout and event handling. This means that long-running JS functions can cause page slowdowns or delays for handling events. To mitigate this issue, JS users rely on event callbacks and promises, essential components of the modern JS language. A prominent example is Node.js which relies on asynchronous I/O and event-driven programming, allowing the development of high-performance, concurrent programs.

%allow developing high-performance, concurrent programs that don't rely on the mainstream multi-threading approach, but on asynchronous I/O and an event-driven programming model.

\begin{figure}[htpb]
\includegraphics[width=3.3in]{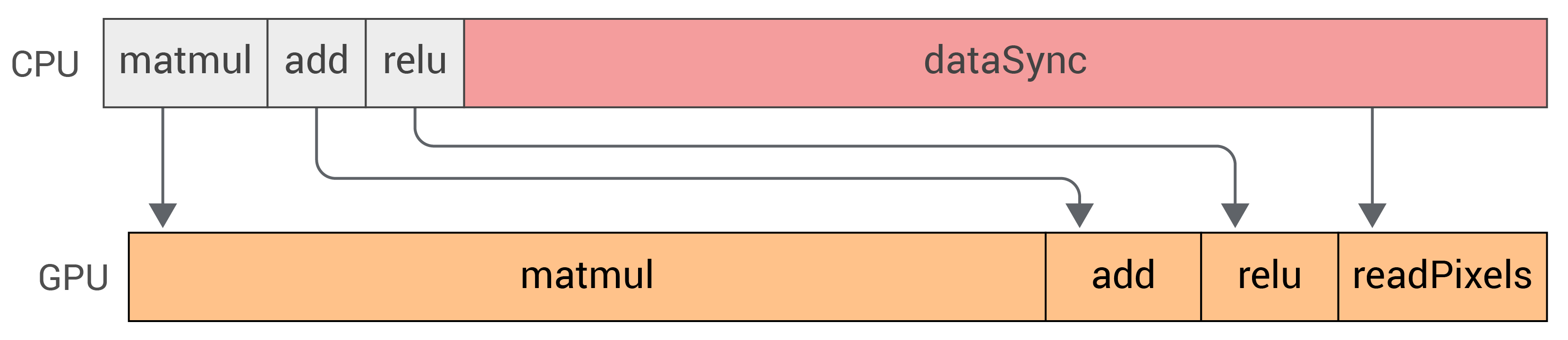}
\caption{The timeline of a synchronous and blocking \textit{tensor.dataSync()} in the browser. The main thread blocks until the GPU is done executing the operations.}
\label{fig:timeline-datasync}
\end{figure}

\begin{figure}[htpb]
\includegraphics[width=3.3in]{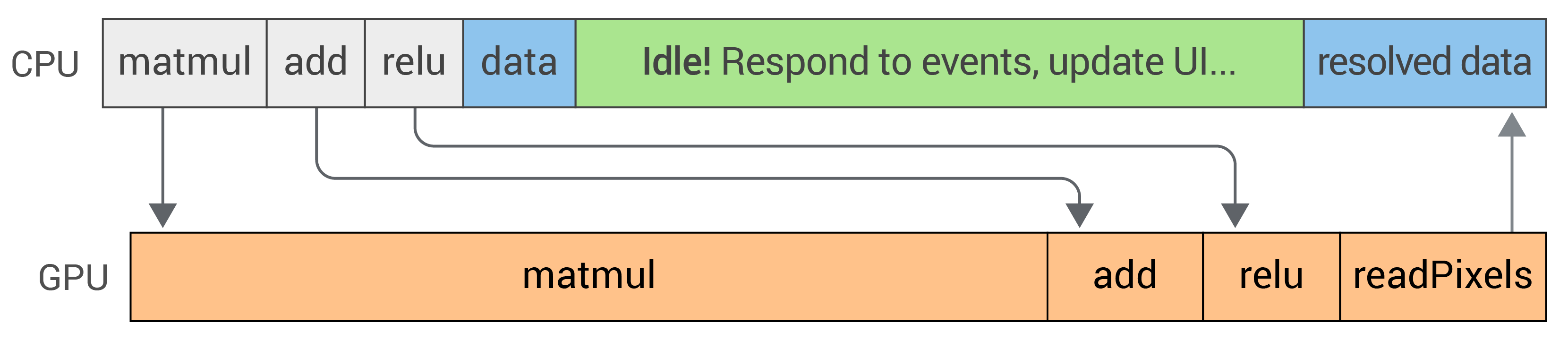}
\caption{The timeline of an asynchronous call to data() in the browser. The main thread is released while the GPU is executing the operations and the data() promise resolves when the tensor is ready and downloaded.}
\label{fig:timeline-data}
\end{figure}

However, callbacks and asynchronous functions can lead to complex code. In service of our design goal to provide intuitive APIs, TensorFlow.js aims to balance the simplicity of synchronous functions with the benefits of asynchronous functions. For example, operations like \textit{tf.matMul()} are purposefully synchronous and return a tensor whose data might not be computed yet. This allows users to write regular synchronous code that is easy to debug. When the user needs to retrieve the data that is backing a tensor, we provide an asynchronous \textit{tensor.data()} function which returns a promise that resolves when the operation is finished. Therefore, the use of asynchronous code can be localized to a single \textit{data()} call. Users also have the option to call \textit{tensor.dataSync()}, which is a blocking call. Figures \ref{fig:timeline-datasync} and \ref{fig:timeline-data} illustrate the timelines in the browser when calling \textit{tensor.dataSync()} and \textit{tensor.data()} respectively.

\subsection{Memory management}

JS provides automatic garbage collection. However, in the browser WebGL memory is not automatically garbage collected. Because of this, and the lack of finalization, we expose an API for all backends to explicitly manage memory.

To dispose the memory allocated by a tensor, users can call \textit{tensor.dispose()}. This approach is relatively straightforward, but the user has to have a reference to all tensor objects so they can be disposed. Often models are written as chained blocks of operations, so breaking up the chains for disposal can be cumbersome. Since tensors are immutable and operations are functional, a single op call can allocate a significant number of intermediate tensors. Forgetting to dispose these intermediate tensor results in memory leaks and slows down the application significantly.

TensorFlow.js offers an alternative approach. Since functions are first-order citizens in JS, and a large portion of the native JS API uses functions as arguments, we decided to provide a scoping mechanism where the user can wrap any synchronous function $f$ by calling \textit{tf.tidy(() }$\Rightarrow$ \textit{f())}. This results in calling $f$ immediately, and disposing all intermediate tensors created inside once $f$ finishes, except for the return result of $f$. We use this mechanism extensively in our library. Users of the Layers API do not need explicit memory management due to model-level APIs such as \textit{model.fit()}, \textit{model.predict()} and \textit{model.evaluate()} which internally manage memory.

\subsection{Debugging and profiling}

TensorFlow.js provides a rich set of debugging tools to help developers understand common problems with performance and numerical stability, accessible either via a URL change or a feature flag. Users can profile every kernel that gets called, seeing the output shape, memory footprint, as well as device-specific timing information. In this mode, every tensor gets downloaded from the GPU and is checked for NaNs, throwing an exception at the first line a NaN is introduced, showing model developers which operation is the source of the numerical instability.

TensorFlow.js also provides \textit{tf.time(f)} for timing a function that calls TensorFlow.js operations. When calling \textit{tf.time(f)}, the function $f$ will be executed and timed. Each backend is responsible for timing functions, as timing may be device specific. For example, the WebGL backend measures the exact GPU time, excluding time for uploading and downloading the data.

A more generic API, \textit{tf.profile(f)}, similarly takes a function $f$ and returns an object representing the function's effect on memory. The object contains the number of newly allocated tensors and bytes created by executing the function, as well as the peak tensors and memory allocated inside the function. Understanding peak memory usage is especially important when running on devices with limited memory such as mobile phones.

\subsection{Performance}

While performance was not the single most important goal, it was critical in enabling real-world ML in JS. In the browser, TensorFlow.js utilizes the GPU using the WebGL API to parallelize computation. By using WebGL for numerical computation, we were able to achieve 2 orders of magnitude speedup, which is what fundamentally enabled running real-world ML models in the browser. On the server-side, TensorFlow.js binds directly to the TensorFlow C API, which takes full advantage of native hardware acceleration.

Table~\ref{tab:perf} shows the speedups of these implementations relative to the plain JS CPU counterpart. We measure a single inference of MobileNet v1 1.0 \cite{mobilenet} with an input image of size \textit{224x224x3}, averaged over 100 runs. All measurements, other than those mentioning GTX 1080, are measured on a MacBook Pro 2014 laptop, while the GTX 1080 measurements are done on a desktop machine. Note that the WebGL and the Node.js CPU backends are two orders of magnitude faster than the plain JS backend, while those utilizing the capable GTX 1080 graphics card are three orders of magnitude faster.

\begin{table}[!htb]
  \begin{center}
    \begin{tabular}{|l r r c|} 
      \hline
      \textbf{Backend} & \textbf{Time (ms)} & \textbf{Speedup} & \\ [0.5ex] 
      \hline
      Plain JS & 3426 & 1x & \\ 
      WebGL (Intel Iris Pro) & 49 & 71x & \\
      WebGL (GTX 1080) & 5 & 685x & \\
      Node.js CPU w/ AVX2 & 87 & 39x & \\
      Node.js CUDA (GTX 1080) & 3 & 1105x & \\ [1ex]
     \hline
    \end{tabular}
  \end{center}
  \caption{Speedups of the WebGL and Node.js backends over the plain JS implementation. The time shows a single inference of MobileNet v1 1.0 \cite{mobilenet}, averaged over 100 runs.}\label{tab:perf}
\end{table}

Since the launch of TensorFlow.js, we have made significant progress on improving our WebGL utilization. One notable improvement is \textit{packing}, where we store floating point values in all 4 channels of a texel (instead of using only 1 channel). Packing resulted in 1.3-1.4x speedup of models such as PoseNet \cite{posenet} across both mobile and desktop devices.

While we will continue to work on our WebGL implementation, we observed a 3-10x gap in performance between WebGL and CUDA. We believe the gap to be due to WebGL's lack of work groups and shared memory access, benefits provided by general-purpose computing (GPGPU) frameworks like CUDA\cite{cuda} and OpenGL Compute shaders \cite{opengl}. As we discuss below in Section~\ref{sec:future-backends}, we believe that the upcoming WebGPU \cite{webgpu} standard is a promising avenue for bridging the gap in performance.

\section{Implementation}\label{sec:impl}

This section describes the specific constraints and implementations of the various backends that are supported by TensorFlow.js.

\subsection{Browser and WebGL}
With the advent of deep learning and scientific computing in general, and advances in modern GPU architectures, the use of GPGPU has grown tremendously. While modern JS virtual machines can optimize plain JS extensively, its performance is far below the computational power that GPUs provide (see Table~\ref{tab:perf}). In order to utilize the GPU, TensorFlow.js uses WebGL, a cross-platform web standard providing low-level 3D graphics APIs. Unlike OpenCL and CUDA, the WebGL API is based on OpenGL ES specification \cite{opengl} which has no explicit support for GPGPU.

Among the three TensorFlow.js backends, the WebGL backend has the highest complexity. This complexity is justified by the fact that it is two orders of magnitude faster than our CPU backend written in plain JS. The realization that WebGL can be re-purposed for numerical computation is what fundamentally enabled running real-world ML models in the browser.

To work around the limitations and the complexities of WebGL, we wrote a layer of abstraction called the \textit{GPGPUContext} which executes WebGL fragment shaders representing computation. In a graphics program, fragment shaders are typically used to generate the colors for the pixels to be rendered on the screen. Fragment shaders run for each pixel independently and in parallel; TensorFlow.js takes advantage of this parallelization to accelerate ML computation.

\begin{figure}[htpb]
\includegraphics[width=3.5in]{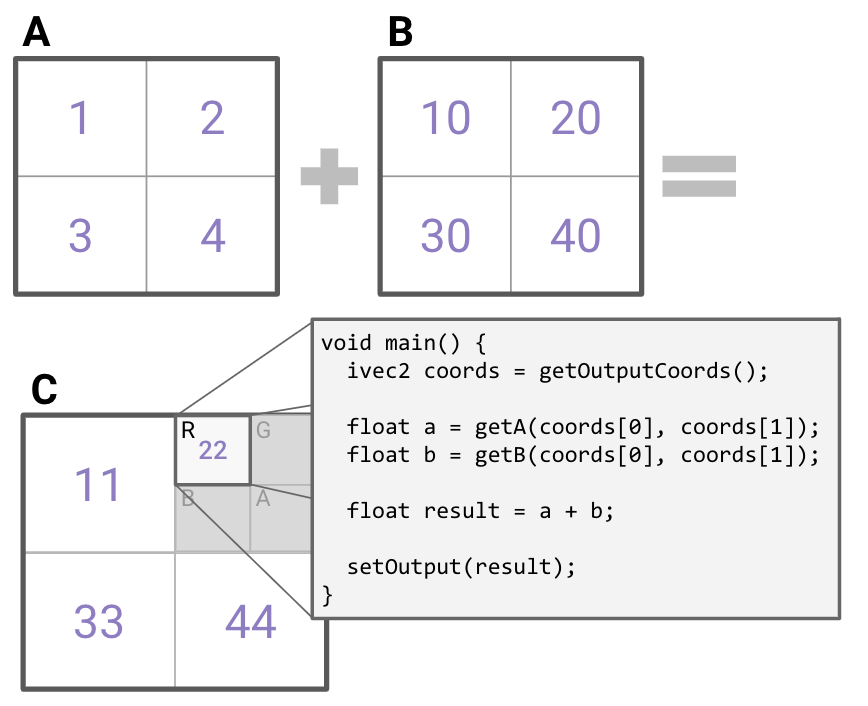}
\caption{The addition of two equally shaped matrices as executed by the WebGL backend, and the GLSL code of the fragment shader that represents the element wise addition computation. The GLSL function, \textit{main()}, runs in the context of each output value and in parallel, with no shared memory.}
\label{fig:fragment-shader}
\end{figure}

In the WebGL backend, the draw pipeline is set up such that the scene geometry represents a unit square. When we execute a fragment shader program, we bind the texture that backs the output tensor to the frame buffer and execute the fragment shader program. This means that the fragment shader \textit{main()} function is executed in parallel for each output value, as shown in ~\ref{fig:fragment-shader}. For simplicity, we only use the red channel of the texture that backs the tensor (shown as `R' in the figure). On WebGL 2.0 devices, we use the \textit{gl.R32F} texture type which allows us to avoid allocating memory for the green, blue, and alpha channels (shown as `G', `B', and `A' respectively). In future work, TensorFlow.js will take advantage of all channels for WebGL 1.0 devices, which will better utilize the GPU's sampler cache.

Writing OpenGL Shading Language (GLSL) code can be error prone and difficult. To make it significantly easier to write and debug GPGPU programs, we wrote a shader compiler. The shader compiler provides high-level GLSL functions that the shader author can call. Listing~\ref{lst:glsl} shows the GLSL source code for matrix multiplication where the shared dimension $N$ is assumed to be a multiple of $4$ for simplicity. The functions marked with bolded font are provided by our shader compiler.

Using the higher level functions generated by the shader compiler has multiple benefits. First, the user-defined GLSL code operates in high-dimensional `logical' space instead of the physical $2D$ texture space. For example, the GLSL implementation of \textit{tf.conv2d()} uses the auto-generated \textit{getA(batch, row, column, depth)} method to sample from a 4D tensor. This makes the user code simpler, more readable and less error-prone.

Second, the separation of logical and physical shape allows the framework to make intelligent decisions about memory layout, avoiding device-specific size limits of WebGL textures.

Third, we can optimize the mapping from high-dimensional space to the $2D$ space. For example, assume the logical shape of tensor $A$ is $4D$ with shape \textit{1x3x1x2}. When $A$ gets uploaded to the GPU, the backend will allocate a physical \textit{3x2} texture and the compiler will generate a $getA(a, b, c, d)$ method whose implementation ignores $a$ and $c$ and directly maps $b$ and $d$ into the $2D$ texture space. We observed that this optimization leads to 1.3x speedup on average.

Last, there is a single GLSL implementation of \textit{tf.matMul()} regardless of the browser's WebGL capabilities. In Chrome we render to a 32bit single-channel floating texture, while in iOS Safari we render to a 16bit single-channel floating point texture. In both cases, the user code is the same, using the high-level \textit{setOutput(value)} GLSL method with the browser-specific implementation generated by the compiler.

\begin{minipage}{\linewidth} % Disable page breaks
\begin{lstlisting}[caption={GLSL code for matrix multiplication using the higher-level utility functions marked in bold font.},label={lst:glsl}]
void main() {
  ivec2 coords = getOutputCoords();
  int aRow = coords.x;
  int bCol = coords.y;
  float result = 0.0;
  for (int i=0; i <${N}; i+=4) {
    vec4 a = vec4(
      getA(aRow, i), getA(aRow, i+1),
      getA(aRow, i+2), getA(aRow, i+3));
    vec4 b = vec4(
      getB(i, bCol), getB(i+1, bCol),
      getB(i+2, bCol), getB(i+3, bCol));
    result += dot(a, b);
  }
  setOutput(result);
}
\end{lstlisting}
\end{minipage}

%The GPGPU context is wrapped by the WebGL backend which implements the linear algebra ops, called kernels. Each call to a kernel in the WebGL backend runs a shader program (left side of Fig.~\ref{fig:stack}). The backend has several performance optimizations such as caching of pre-compiled shader binaries, as well as a custom texture manager that keys textures based on $2D$ shape, and re-uses them upon future allocation.

\subsubsection{Asynchronous execution}

With WebGL, programs get scheduled by the CPU and run on the GPU, which is a separate thread from the main JS thread. This means that while programs are running on the GPU, the CPU is free to respond to events and run other JS code.

When the user calls an operation, we enqueue a program onto the GPU command queue, which typically takes sub-millisecond time, and immediately return a handle to the resulting tensor despite the computation not being done. Users can later retrieve the actual data by calling \textit{tensor.dataSync()} or \textit{tensor.data()}, which returns a \textit{TypedArray}.

%The synchronous \textit{tensor.dataSync()} method calls the underlying WebGL \textit{gl.readPixels()} function and returns the values of the tensor. However, since the \textit{gl.readPixels()} call is synchronous, it blocks the main thread as shown in Fig.~\ref{fig:timeline-datasync}.

As mentioned in Section~\ref{sec:async}, we encourage the use of the asynchronous \textit{tensor.data()} method, which avoids blocking the main thread, and returns a promise that resolves when the computation is done (see Figures~\ref{fig:timeline-datasync} and \ref{fig:timeline-data}). However, to retrieve the underlying data of a texture, the WebGL API only provides a blocking \textit{gl.readPixels()} method. To get around this limitation, we approximate when the GPU is done executing the operations, postponing the call to \textit{gl.readPixels()}, which releases the main thread in the meantime.

Approximating when the GPU has finished executing programs can be done in a couple of ways. The first approach taken in TensorFlow.js, for WebGL 1.0 devices, uses the \textit{EXT\_disjoint\_timer\_query} WebGL extension. This extension can be used to accurately measure the GPU time of programs, but also implicitly has a bit that gets flipped when a program is done executing. The second approach, for WebGL 2.0 devices, uses the \textit{gl.fenceSync()} API by inserting a `fence' into the GPU command queue and polling a query which returns true when the fence has flipped.

% To transfer data from the GPU to the CPU, we call \code{gl.readPixels} on a texture which resolves with the values when the program that generates the texture has finished.

%In TensorFlow.js, kernels are expressed as WebGL programs and tensors are backed by WebGL textures. After a user has called an operation and wants to get the result from the operation, we must provide a way to synchronize the CPU and the GPU and download values off of the GPU into a CPU ArrayBuffer. , which means other events or callbacks can never fire while we are performing linear algebra on the GPU.
%To mitigate this problem, users should call the asynchronous \textit{tensor.data()} method, which returns a \textit{promise} that resolves with the underlying values. Though native WebGL only provides a synchronous \textit{gl.readPixels()} call, we can approximate when the GPU is done executing the operations, releasing the UI thread until it is done and only then calling the blocking \textit{gl.readPixels()} call, which can be seen in Fig.~\ref{fig:timeline-data}.

\subsubsection{Memory management}

Disposing and re-allocating WebGL textures is relatively expensive, so we don't release memory when a tensor gets disposed. Instead, we mark the texture for reuse. If another tensor gets allocated with the same physical texture shape, we simply recycle the texture. The texture recycler gives us significant performance wins since multiple passes through the same ML model often generate tensors of the same shapes. 

One of the common problems with manual memory management is memory leaks, where a program with a loop creates one or more tensors during each tick that never get disposed. This will eventually cause the application to run out of memory and crash. Since one of our primary design principles is easy-of-use over performance, we provide built-in heuristics to avoid crashing the application. Specifically, we automatically page WebGL textures to the CPU when the total amount of GPU memory allocated exceeds a threshold which can be estimated from the screen size. At the same time, the paging logic will not take effect for users that explicitly manage memory using \textit{tf.tidy()} or \textit{tensor.dispose()}.

\subsubsection{Device support}
While WebGL is a slow-evolving web standard, it has ubiquitous support. TensorFlow.js can run on desktop, tablet and mobile devices. The WebGL backend requires a WebGL 1.0 compatible device that supports the \textit{OES\_texture\_float} extension which enables uploading and reading from floating point textures. According to WebGLStats.com \cite{webglstats}, a website that tracks the WebGL capabilities of devices and their market share on the web, TensorFlow.js can run on 99\% of desktop devices, 98\% of iOS and Windows mobile devices, and 52\% of Android devices. We believe that the reason for the significant Android discrepancy is due to a large number of older Android devices that have no GPU hardware, and that this gap will gradually close as users migrate to newer Android devices.

While WebGL has wide support on different platforms, not all browsers have the exact same implementation of the WebGL specification. This leads to problems with the numerical stability of the library. For example, on iOS devices, the WebGL API is explicit about the lack of 32bit float support, reverting to 16bit floats, which is aligned with the underlying capability of mobile GPUs. On the other hand, mobile Chrome hides that detail by allowing developers to upload and write in 32bit float regardless of the underlying precision. This led to numerical precision problems on Android devices: $log(x+\epsilon)$ resulted in $log(x+0)$ since the default $\epsilon=\num{1e-8}$ was not representable in 16bit float and implicitly rounded to $0$. To solve this problem, we adjust the global $\epsilon$ value according to the device capabilities.
%we explicitly test for 16bit support by running \code{tf.abs(\num{1e-5})} in a shader, checking if the result is 0 and adjusting the global $\epsilon$ value accordingly.

\subsection{Node.js}
With the advent of Node.js and event-driven programming, the use of JS in server-side applications has been steadily growing \cite{nodejs}. While the browser as an execution platform has significant limitations, server-side JS has full access to the filesystem, native operating system kernels, and existing C and C++ libraries.

To support the use-case of server-side ML in JS, we provide a Node.js backend that binds to the official TensorFlow C API using the N-API \cite{napi}. While the internal backend has a different implementation than the WebGL backend, they share the same user-facing API, enabling full portability between the server and the web browser.

The Node.js backend has distinct advantages. Since Node.js and Google's V8 JS engine exposes finalization APIs, it eliminates the need for manual memory management, reducing the cognitive overhead for our users. Binding to the TensorFlow C library means full advantage of the flexibility and performance of TensorFlow. Under the hood, we utilize hardware acceleration both on the CPU, with AVX instructions, and the GPU with the CUDA and CuDNN libraries \cite{cuda}. Binding to the TensorFlow C API means that TensorFlow.js will support TPUs (Tensor Processing Units) in a future release of the Node.js binding.

\subsection{Future backends} \label{sec:future-backends}

Two new web standards, WebAssembly and WebGPU, have potential to improve TensorFlow.js performance.

WebAssembly is a binary instruction format for the web, designed as a potential target for compilation of languages like C and C++. At present, WebAssembly is enabled in most major browsers. By shipping lower-level instructions, WebAssembly compiled from C can see performance gains over vanilla JS. Moreover, WebAssembly will allow writing SIMD code in C, which speeds up linear algebra by computing dot products of vectors in a single instruction. Currently, WebAssembly does not support SIMD.

WebGPU is the working name for a future web standard and JS API for accelerated graphics and compute. WebGPU provides a more generic way to express parallelizable computation on the GPU, which would allow us to write more optimized linear algebra kernels than the ones with the WebGL backend.

\section{Ecosystem Integration}

This section describes how TensorFlow.js fits into the broader TensorFlow ecosystem.

\subsection{Model Converter}
\label{sec:converter}
%The ability to author models in TensorFlow.js plays a vital role in bringing ML to JS developers, by allowing users to create and tweak existing models, all in JS. 
There is a plethora of pre-trained, open-sourced models, developed in TensorFlow, that are targeted for edge devices. TensorFlow.js offers a model converter that can load and execute pre-trained TensorFlow SavedModels and Keras models, allowing these models to be available in JS.

To port an existing model to TensorFlow.js, the user runs a Python script that converts the existing format to the TensorFlow.js web format. TensorFlow.js optimizes the model by pruning unnecessary operations (e.g. training operations) and packs weights into 4MB files, optimizing for browser auto-caching. 
%During conversion, we run a graph optimizer that prunes the graph from unnecessarily operations (e.g. training operations) and packs weights into 4MB files, optimizing for browser auto-caching.
The user can also quantize the weights, reducing the model size by 4X. After the conversion, in the JS application the user can call \textit{tf.loadModel(url)} to load the model, providing a URL where the web files are hosted.

\subsection{Models repo}

One of the major benefits of the JS ecosystem is the ease at which JS code and static resources can be shared. TensorFlow.js takes advantage of this by hosting an official repository of useful pretrained models, serving the weights on a publicly available Google Cloud Storage bucket. This makes it easy for beginners to integrate these models into their existing applications.

Furthermore, to address our core goal of enabling ML beginners, we provide wrapper APIs that hide tensors from the user. The model prediction methods always take native JS objects like DOM elements or primitive arrays and return JS objects that represent human-friendly predictions. Listing ~\ref{lst:posenet-api} shows an example application using PoseNet \cite{posenet}, a model hosted in the repository that computes human pose estimations from an image. Note that the user does not need to use \textit{tf.Tensor} to use the PoseNet model.

One of our core design principles is not to sacrifice functionality for simplicity. For these reasons, we expose APIs to work with tensors for expert users. For these users, these models can be used in a transfer learning setting, enabling personalized applications with on-device training with relatively little user data.

\begin{minipage}{\linewidth} % Disable page breaks
\begin{lstlisting}[mathescape=true, caption={An example showing the PoseNet API which allows passing an \textit{HTMLImageElement} to the pose estimate method and returns a JS object with the predictions. },label={lst:posenet-api}]
const imageElement =
    document.getElementById('person');

// Estimate a single pose from the image.
posenet.estimateSinglePose(imageElement)
    .then(pose => console.log(pose));

$\mbox{\textbf{Console output:}}$
{
  "score": 0.32,
  "keypoints": [
    {
      "position": {"x": 253.37, "y": 76.29},
      "part": "nose",
      "score": 0.99
    },
    ...
  ]
}
\end{lstlisting}
\end{minipage}

\section{Examples and Usage}

Since launching TensorFlow.js publicly in March 2018, we have seen excitement about TensorFlow.js in a few different domains which we outline in this section.

\subsection{Education and Learning}

The in-browser implementation of TensorFlow makes it easy to get started with \cite{carrie}. Educators can deploy interactive lessons to students without the additional burden of software installation. One early success we saw with TensorFlow.js in the educational deep learning space was Teachable Machine \cite{google_teachable_2017}, an application that allows visitors to build their own image classifier directly in the browser using their webcam, no coding required. The site saw over 450,000 unique visits, and the GitHub code has over 3000 stars. Another successful education application is GAN Lab \cite{kahng_gan_2018} \footnote{https://poloclub.github.io/ganlab/}, a tool that enables interactive visual exploration of Generative Adversarial Networks (GANs). Targeted towards non-experts, it helps users develop a greater intuitive sense of how GANs work during training and inference.

TensorFlow.js extends the deep learning ecosystem to the JS communities that are unfamiliar with Python and C. Dan Shiffman, at NYU’s Interactive Telecommunications Program, has made both an online video course on ML with JS, and a JS ML library called ML5 \cite{noauthor_ml5js_nodate}, built on top of TensorFlow.js. ML5 aims to provide `friendly ML for the web' and is geared towards artists, creative coders and students, empowering them to innovate in new ways. The ML5 team identified installation issues as the first barrier that beginners face when trying to approach ML \cite{shiffman_ml5:_2018} and built upon TensorFlow.js to overcome that barrier.

%TensorFlow.js and ML5 significantly reduce that barrier allowing learners to start in environments that require no installation. 

The Deep Learning Practicum at MIT \cite{abelson_6.s198_2018} is another example of the value of TensorFlow.js in an educational setting. This course provides undergraduate students with hands-on experience building deep learning models in JS while they learn the theory and concepts behind the algorithms they are working with.

We also see a great deal of self-directed learning and exploration using TensorFlow.js. Many people who had a passing interest in ML, but found it difficult to access, use TensorFlow.js as an opportunity to learn about the new technology with fewer barriers to entry. This includes people building simple demos and training their own object recognizers, to those exploring cutting edge topics such as NeuroEvolution \cite{thebe_evolutionsimulator:_2018} and Reinforcement Learning \cite{neveu_metacar:_2018}. These demos were built by the authors to develop their ML skills and also to allow others to easily experiment without the infrastructure typically associated with these types of algorithms. These use cases demonstrate a trade-off between accessibility and performance where a bias towards accessibility is acceptable and even embraced.
%that are acceptable, and are even embraced for certain classes of applications.
Many more examples of this type of application have been built and are accessible through our community project gallery  \footnote{https://github.com/tensorflow/tfjs/blob/master/GALLERY.md}: Simple MNIST GAN \cite{chang_mnist_nodate}, Emotion Extractor \cite{sudol_emotion_nodate}, Next Word Predictor \cite{malviya_next_nodate} and more.

\subsection{Gestural Interfaces}

Real time applications that make use of gestural inputs with the webcam is another area where TensorFlow.js has seen promising results. TensorFlow.js users have built applications that enable sign language to speech translation \cite{singh_alexa-sign-language-translator:_2018}, enable individuals with limited motor ability control a web browser with their face \cite{ramos_handsfree.js_nodate}, and perform real-time facial recognition and pose-detection \cite{friedhoff_move_2018}. Each of these examples is powered by a pre-trained image model, usually MobileNet \cite{mobilenet}, that is fine-tuned for the project, or expose interactive fine-tuning interfaces to the end user.

\subsection{Research Dissemination}

TensorFlow.js has also enabled ML researchers to make their algorithms more accessible to others. For example, the Magenta.js \cite{magentajs} library provides in-browser access to generative music models developed by the Magenta team and ported to the web with TensorFlow.js. Magenta.js has increased the visibility of their work with their target audience, namely musicians. This has unleashed a wide variety of ML powered music apps built by the community such as Latent Cycles \cite{parviainen_latent_nodate} and Neural Drum Machine \cite{parviainen_neural_nodate}. These and more examples can be found at https://magenta.tensorflow.org/demos.

We have also seen TensorFlow.js used to power interactive client-side ML in web-based scholarly articles \cite{ha_world_2018} \cite{carter_using_2017}, offering a richer communication medium where dynamic behaviour of models can be demonstrated in real time and manipulated by the reader.

\subsection{Numeric Applications}

Another category of applications that TensorFlow.js enables is GPU accelerated tools for numerical computation. An example is tfjs-tsne \cite{pezzotti2018linear}, a novel linear time approximation of the t-SNE algorithm that runs in the browser. TensorFlow.js's GPU acceleration primitives make it practical to run it interactively in the browser for datasets of tens of thousands of points.

\subsection{Desktop and Production Applications}

An emerging area where JS has been applied is in desktop and production applications, demonstrating the wide reach of the JS ecosystem.

Node Clinic is an open source Node.js performance profiling tool that recently integrated a TensorFlow.js model to separate CPU usage spikes caused by the user from those caused by Node.js internals (e.g. garbage collection) \cite{nearform_node_nodate}.

\textit{Mood.gg Desktop} \cite{farza_deepoverwatch_2018}, created by a student, is a desktop application powered by Electron, a popular framework for writing cross-platform desktop apps in JS. It uses a TensorFlow.js model trained to detect which character the user is playing in a popular team based game called Overwatch, by looking at the user's screen. This is used to play a custom soundtrack from a music streaming site, Mood.gg, that matches the music to the playing style of the in-game character (e.g. `death metal' for the character called `Reaper'). The character prediction from the pixels of the screen happens entirely client-side, preserving the privacy of the player and highlighting a key advantage of client-side ML. The author reports that over 500,000 people use the site.

\section{Conclusion and future work}

TensorFlow.js is a high-performance deep learning toolkit in JS that runs both on the client and the server. It is an accessible on-ramp for deep learning to a community that often focuses on the end user. As such, TensorFlow.js has the potential to greatly broaden the set of people who can take advantage of modern ML techniques. We have already seen a rich variety of applications of TensorFlow.js.

A key technical contribution of TensorFlow.js is the set of techniques used to repurpose the web platform's graphics APIs for high-performance numeric computing while maintaining compatibility with a large number of devices and execution environments.

% I would add something about the API here.

We believe there are a number of opportunities to extend and enhance TensorFlow.js. Given the rapid progress of browser development, it seems likely that additional GPU programming models may become available. In particular, we see conversations by browser vendors to implement general purpose GPU programming APIs \cite{apple_next-generation_2017} \cite{w3gpu_nodate} that will make these kinds of toolkits more performant and easier to maintain.

%Looking forward we are excited about a number of things: first, the web platform is only going to get better. 10 years ago many wouldn't have imagined the ubiquity and richness of highly interactive web applications we take for granted today. Browsers will continue to gain more and more capabilities as they are the premiere platform for mass distribution of end-user computer applications. Specifically we see movement around conversations by browser vendors to implement general purpose GPU programming APIs \cite{apple_next-generation_2017} \cite{w3gpu_nodate} that will make these kinds of toolkits more performant and easier to maintain.

Future work will focus on improving performance, continued progress on device compatibility (particularly mobile devices), and increasing parity with the Python TensorFlow implementation. We also see a need to provide support for full machine learning workflows, including data input, output, and transformation. More generally, we see a broad opportunity for contributing to the burgeoning JS data science ecosystem \cite{manning-book-data-js}, with the goal of decreasing the difficulty of ML development, increasing participation in the ML community, and allowing new types of applications.

%creating data pipelines to get data in and out of ML models and will be working on APIs to support that and provide easy access to best practices such as normalization and shuffling.

% Acknowledgements should only appear in the accepted version.
%\section*{Acknowledgements}
%Make sure to thank a bunch of people and oss contributors.

% Mahima, Manraj, Lewauthe, Josh Gartman

% In the unusual situation where you want a paper to appear in the
% references without citing it in the main text, use \nocite
%\nocite{langley00}

\bibliography{references}
\bibliographystyle{sysml2019}

\end{document}